\documentclass{article}
\usepackage{ijcai25}

\usepackage{times}
\usepackage{soul}
\usepackage{url}
\usepackage[hidelinks]{hyperref}
\usepackage[utf8]{inputenc}
\usepackage[small]{caption}
\usepackage{graphicx}
\usepackage{amsmath}
\usepackage{amsthm}
\usepackage{booktabs}
\usepackage{algorithm}
\usepackage{algorithmic}
\usepackage[switch]{lineno}
\usepackage{amssymb}
\usepackage{multirow}
\usepackage{tabularx} 

\title{CLCR: Contrastive Learning-based Constraint Reordering for\\ Efficient MILP Solving}

\author{
Shuli Zeng
\and
Mengjie Zhou\and
Sijia Zhang\and
Yixiang Hu\and
Feng Wu\And
Xiang-Yang Li$^*$\\
\affiliations
School of Computer Science and Technology, University of Science and Technology of China\\
\emails
\{zengshuli0130,mjzhou,sxzsj,yixianghu\}@mail.ustc.edu.cn,
\{wufeng02,xiangyangli\}@ustc.edu.cn
}

\begin{document}

\maketitle

\begin{abstract}
Constraint ordering plays a critical role in the efficiency of Mixed-Integer Linear Programming (MILP) solvers, particularly for large-scale problems where poorly ordered constraints trigger increased LP iterations and suboptimal search trajectories. This paper introduces CLCR (Contrastive Learning-based Constraint Reordering), a novel framework that systematically optimizes constraint ordering to accelerate MILP solving. CLCR first clusters constraints based on their structural patterns and then employs contrastive learning with a pointer network to optimize their sequence, preserving problem equivalence while improving solver efficiency. Experiments on benchmarks show CLCR reduces solving time by 30\% and LP iterations by 25\% on average, without sacrificing solution accuracy. This work demonstrates the potential of data-driven constraint ordering to enhance optimization models, offering a new paradigm for bridging mathematical programming with machine learning.

\end{abstract}

\section{Introduction}

Mixed-integer linear programming (MILP), which seeks to optimize a linear objective function under both linear constraints and integer constraints (i.e., some or all decision variables are restricted to integer values), is a cornerstone in the field of mathematical optimization ~\cite{DBLP:books/siam/04/BixbyFGRW04,DBLP:phd/de/Achterberg2007}. Due to its expressiveness and versatility, MILP has been widely adopted to model and solve a variety of real-world optimization problems, including supply chain management~\cite{paschos2014applications}, production planning~\cite{junger200950}, scheduling~\cite{chen2010integrated}, vehicle routing~\cite{laporte2009fifty}, facility location~\cite{farahani2009facility}, and bin packing~\cite{nair2020solving}. These applications span critical industries and domains, highlighting the practical importance of MILP in decision-making and operational efficiency.

Despite its extensive applicability, solving MILP instances remains inherently challenging due to their classification as NP-hard problems~\cite{DBLP:books/siam/04/BixbyFGRW04}. Consequently, the development of efficient algorithms for tackling MILP has become a focal point of research in both academia and industry. Recent years have witnessed a growing interest in leveraging machine learning (ML) techniques to enhance the performance of traditional combinatorial optimization solvers. These ML-based approaches hold promise, particularly when dealing with problems arising from specific distributions, thanks to their capacity to learn and adapt to recurring patterns and inherent structures that are often found in real-world scenarios~\cite{zhang2023survey}. Traditional hand-crafted rules struggle to capture these subtle variations effectively~\cite{li2022learning}.

Typically, ML methods have focused on augmenting or replacing specific solver components, such as cutting plane selection~\cite{wang2023learning} and node selection~\cite{zhanglearning}. However, relatively little attention has been given to the initial stages of solver performance, specifically during the formulation phase when real-world optimization problems are translated into mathematical programming models. Historically, this translation has been done by human experts who create effective formulations for the problems at hand. Such formulations are then passed to solvers to derive optimal solutions, often without considering how the formulation itself impacts solver efficiency. Recent research in optimization theory suggests that these formulation details, particularly the order of variables and constraints, can significantly influence solver performance~\cite{ge2021interior,li2022learning}.

The order of variables and constraints indeed plays a crucial role, as it influences several solver processes like presolving and linear algebra operations. A well-chosen order can enhance the solver's ability to simplify the problem, efficiently navigate the solution space, and thereby reduce computational costs. For instance, during the presolve phase, solvers rely on the order of constraints to detect redundancies and tighten bounds, which can speed up the problem-solving process. Similarly, for linear algebra operations like those in the Simplex method, the initial ordering affects the efficiency and number of iterations required to reach an optimal solution~\cite{bixby1992implementing,maros2002computational,li2022learning}. Consequently, inefficient ordering can severely hamper performance, particularly in large-scale problems, leading to increased computational burdens.

In this paper, we introduce a novel approach called CLCR (Constraint Learning and Clustering Reordering) that leverages machine learning to optimize constraint ordering, thereby improving solver performance. CLCR employs a data-driven strategy, applying clustering algorithms to categorize constraints, followed by a pointer network to determine the optimal order within these categories. By adjusting the constraint order without altering the MILP's mathematical equivalence, our approach offers a significant advancement in enhancing solver efficiency.

The contributions of this work are summarized as follows:

\begin{itemize}
    \item We propose the first method combining constraint clustering and contrastive learning to reorder MILP constraints. CLCR preserves problem equivalence while accelerating solvers through learned structural patterns.
    
    \item We demonstrate that our approach, CLCR, significantly enhances solver efficiency, achieving up to 30\% reduction in solving time and 25\% fewer LP iterations across various benchmark datasets.
    
    \item This work provides a general-purpose method for improving MILP efficiency by optimizing internal constraint orderings, without relying on enhancements to solver components.
\end{itemize}
\section{Preliminaries} 
\label{sec:background}

\subsection{Mixed Integer Linear Programming}

The general form of a MILP problem is given by
\begin{equation}
z^* = \min_{\mathbf{x}} \left\{\mathbf{c}^T \mathbf{x} \mid \mathbf{A} \mathbf{x} \leq \mathbf{b}, \mathbf{x} \in \mathbb{R}^n, x_j \in \mathbb{Z}, \forall j \in \mathcal{I} \right\}
\label{def_milp}
\end{equation}
where $\mathbf{x} \in \mathbb{R}^{n}$ is the set of $n$ decision variables, $\mathbf{c} \in \mathbb{R}^n$ formulates the linear objective function, $\mathbf{A} \in \mathbb{R}^{m \times n}$ and $\mathbf{b} \in \mathbb{R}^{m}$ formulate the set of $m$ constraints, with $\mathbf{A}$ representing the constraint matrix and $\mathbf{b}$ the vector of constraint bounds. Here, $x_j$ is the $j$-th component of vector $\mathbf{x}$, and $\mathcal{I} \subseteq \{1, \dots, n\}$ is the set of indices for integer-constrained variables. The goal is to find $z^*$, the optimal value of the objective function in Eq~\ref{def_milp}, which respects all given constraints.

\subsection{Core Mechanisms of MILP Solvers}

MILP solvers employ sophisticated techniques to address the complexity of combining integer and continuous decision variables. These techniques are structured around two key stages: presolving and branch-and-bound~\cite{boyd2007branch}.

\textbf{Presolve Phase:}  
The presolve phase reduces the problem complexity by simplifying the model through redundancy elimination, bound tightening, and infeasibility detection. Additionally, heuristic algorithms are used to identify feasible solutions early, which can guide subsequent solving stages.

\textbf{Branch-and-Bound Framework:}  
In the branch-and-bound framework, the MILP problem is divided into smaller subproblems, each solved at a node of the search tree. Continuous relaxations of these subproblems provide bounds for pruning and branching decisions. Key aspects, such as variable and node selection strategies, are guided by heuristic algorithms, which play a crucial role in navigating the search tree efficiently. These heuristics help prioritize promising branches and reduce the computational burden by focusing on subproblems with higher potential for improving the solution.

\textbf{Solving LP Relaxations:}  
At each node, the solver tackles an LP relaxation, temporarily ignoring integer constraints. Algorithms like Simplex~\cite{ficken2015simplex} or Interior Point~\cite{darvay2003new} efficiently solve these relaxations, providing bounds for the branch-and-bound process. These LP solutions are pivotal in determining the solver's progress and ensuring computational efficiency.



\section{Related Work}

\subsection{Reformulating Linear Programming}

Recent advances in machine learning (ML) have introduced novel approaches to reformulating linear programming (LP) problems to enhance solver efficiency. Ge et al.~\cite{ge2021interior} made a significant contribution to this area by addressing the crossover phase, which transitions suboptimal solutions from interior-point methods to optimal basic feasible solutions (BFS). They proposed a tree-based method for minimum-cost flow problems and a perturbation-based crossover for general LPs. By exploiting problem structures, such as network characteristics, their methods identified optimal or near-optimal bases efficiently, significantly reducing computational overhead. Building on this foundation, Li et al.~\cite{li2022learning} explored ML techniques to optimize variable ordering in LP problems. They employed graph convolutional neural networks (GCNN) and pointer networks to learn optimal variable orderings, achieving notable reductions in solving time and iteration counts across diverse datasets. This approach demonstrated the potential of data-driven methods in improving solver performance at the formulation stage. Fan et al.~\cite{fan2023smart} further extended these ML-based reformulation techniques by introducing a method for selecting smart initial bases in the simplex method. By leveraging graph neural networks (GNNs) to predict initial bases, their approach effectively utilized problem similarities across instances. This learning-based strategy showed consistent speedups over traditional rule-based methods and minimized simplex iterations in solvers such as HiGHS and OptVerse. Additionally, they adapted this framework for column generation methods, achieving significant improvements in restricted master problems.

\subsection{Learning Order in Branch-and-Bound}

In MILP solvers, the addition and order of constraints, including cutting planes, play a pivotal role in solver efficiency. Recent research has begun to address the impact of cut selection and their sequential ordering on optimization performance. Wang et al.~\cite{wang2023learning} introduced a hierarchical sequence model (HEM) to tackle the challenges of learning cut selection policies. The importance of order was further highlighted in empirical studies where randomly permuting the same set of cuts resulted in widely varying solver performance. HEM addressed this by using a two-level policy: a higher-level model predicts the number of cuts to be selected, while a lower-level sequence model determines the optimal order.
\section{Motivation} \label{sec:motivation}

\subsection{Order Matters}

The performance of solvers in Mixed-Integer Linear Programming (MILP) problems is highly sensitive to the order in which variables and constraints are presented. The order of these elements can significantly influence the solver's ability to simplify the problem, search the solution space effectively, and perform linear algebra operations efficiently. In this section, we highlight three key stages of the solver process where the order plays a critical role.

\textbf{Presolving Phase:}  
The presolve phase aims to reduce the problem size and eliminate irrelevant parts of the search space by simplifying the MILP model. During this phase, solvers detect redundancies, tighten variable bounds, and identify easy-to-eliminate constraints. A poor ordering of variables and constraints can hinder these simplifications, leading to higher computational costs. Conversely, an improved ordering allows the solver to identify redundant constraints and variables more efficiently, accelerating the presolve process and reducing the overall problem size.

\textbf{Heuristic Algorithms Efficiency:}  
MILP solvers often employ heuristic algorithms to find initial solutions and guide decision-making strategies within the branch-and-bound framework. These heuristic methods, such as large neighborhood search (LNS)~\cite{pisinger2019large}, which relax a subset of constraints and solve the resulting problem to identify feasible solutions, and pseudo-cost-based variable selection strategies that score variables based on their impact, are both influenced by the ordering of constraints and variables~\cite{refalo2004impact}. Different orderings may lead to distinct initial feasible solutions, as well as variations in the search paths and efficiency within the branch-and-bound framework. Such differences can have a profound impact on the overall efficiency of the solver.

\textbf{Linear Program Operations:}  
The order of variables and constraints can also impact the efficiency of linear program operations, particularly in methods such as the Simplex algorithm. For instance, the initial basis selection during the Simplex method depends on the ordering of the variables and constraints, which can affect the number of pivot operations required to reach an optimal solution. Studies have shown that the choice of variable and constraint order significantly influences the efficiency of the LP solver by affecting the computational complexity of the pivoting process~\cite{bixby1992implementing,maros2002computational,li2022learning}. A well-chosen ordering can reduce the number of iterations and pivot operations, leading to faster convergence in the Simplex method and other linear programming solvers.

\subsection{Motivation Results}

\begin{figure}
    \centering
    \includegraphics[width=\linewidth]{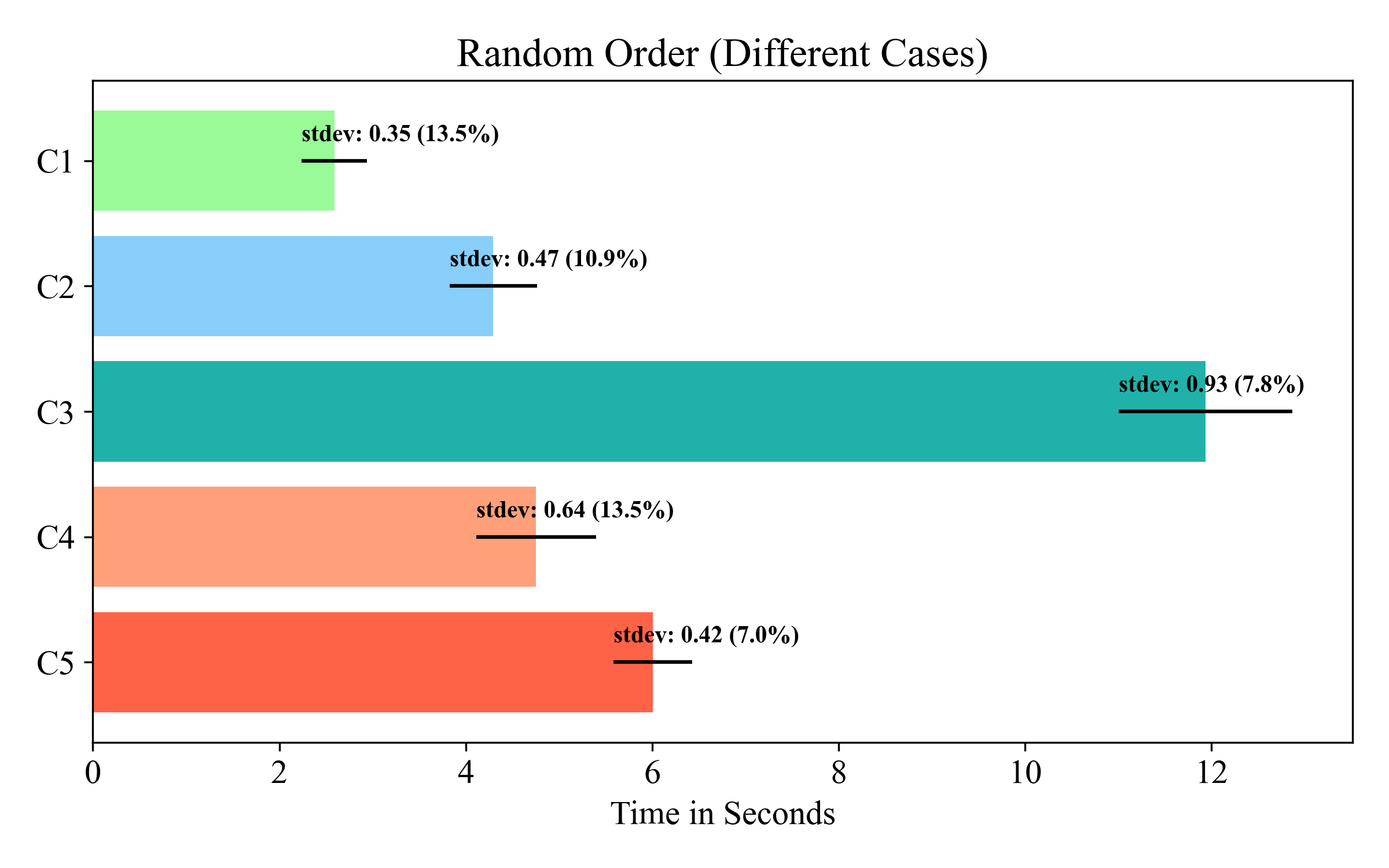}
    \caption{We evaluate the impact of random constraint reordering on solver performance using five different cases. Each bar represents the solver time across 100 random seeds, with the standard deviation and percentage of variation indicated.}
    \label{fig:random_ordering}
\end{figure}

We tested the impact of random constraint reordering on solver performance across different datasets. For each of the five selected cases from Set Cover, we utilized SCIP 8.0.4~\cite{scip} as the backend solver and applied a random permutation to all the constraints and recorded the resulting solver times. As shown in Figure~\ref{fig:random_ordering}, the results highlight the variability in solver performance due to different constraint orders. Each bar represents the average solver time across 100 random seeds, accompanied by the standard deviation (stdev) and percentage of variation. The results indicate that models with different constraint orders exhibit variability in solver performance. This variability underscores the importance of the input order in solving MILP problems, highlighting that the arrangement of constraints can have a significant impact on the solver’s efficiency.

\section{Proposed Method} \label{sec:method}
\begin{figure*}[t]
    \centering
    \includegraphics[width=\linewidth]{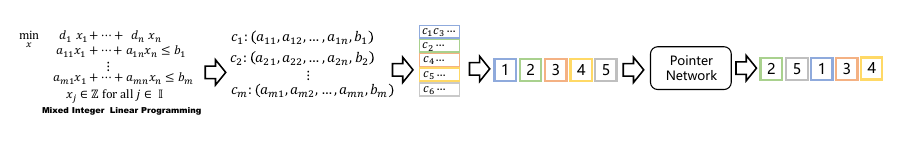}
    \caption{Illustration of the constraint representation, clustering, and reordering process. Constraints are represented as feature vectors and grouped into clusters using K-means. A pointer network is then used to determine the optimal ordering of the clusters, which is applied to reorder the constraints in the original MILP problem.}
    \label{fig:reorder}
\end{figure*}

In this section, we introduce our constraint reordering process and the contrastive learning-based training approach. We represent the constraints of a given MILP problem, cluster them using a clustering algorithm, and extract features from each group. These features are processed by a pointer network to generate a new constraint order, which is used to reformulate the original MILP problem for improved solver performance. Contrastive learning is employed to enhance the representation learning process. The workflow is illustrated in Figure~\ref{fig:reorder}.

\subsection{Clustering and Representing Constraints}

We aim to represent and cluster constraints in MILP problems to uncover structural patterns and simplify downstream processing. Each constraint is described by its coefficients $\mathbf{a} = [a_1, a_2, \dots, a_n]$ and right-hand side (RHS) value $b$, forming a feature vector $\mathbf{f} = [\mathbf{a}, b]$ that captures its key properties.

To group similar constraints, we use the K-means clustering algorithm~\cite{macqueen1967some}, a popular unsupervised learning method. This algorithm partitions the constraints into $k$ clusters by minimizing the sum of squared Euclidean distances between each feature vector and the centroid of its assigned cluster. The optimization objective of K-means is:
\begin{equation}
\min_{\{\mathbf{C}_i\}_{i=1}^k} \sum_{i=1}^k \sum_{\mathbf{f} \in \mathbf{C}_i} \|\mathbf{f} - \boldsymbol{\mu}_i\|^2,
\end{equation}
where $\mathbf{C}_i$ denotes the $i$-th cluster, $\boldsymbol{\mu}_i$ is its centroid, and $\|\cdot\|$ represents the Euclidean norm. 

After clustering the constraints, we extract features for each cluster to summarize its overall characteristics. Specifically, we employ average pooling to aggregate the features of all constraints within a cluster. For a cluster $\mathcal{C}_i$ containing $n_i$ constraints with feature vectors $\mathbf{f}_1, \mathbf{f}_2, \dots, \mathbf{f}_{n_i}$, the cluster-level feature vector $\mathbf{F}_i$ is computed as:
\begin{equation}
\mathbf{F}_i = \frac{1}{n_i} \sum_{j=1}^{n_i} \mathbf{f}_j,
\end{equation}
where $\mathbf{F}_i$ represents the average feature vector for the $i$-th cluster. This aggregation captures the collective properties of the constraints within each cluster, providing a concise and informative representation.

\subsection{Pointer Network-Based Reordering}

Once the cluster features are obtained, we utilize a pointer network~\cite{vinyals2015pointer} to determine the optimal ordering of the clusters. The pointer network is a sequence-to-sequence model that has been widely used in combinatorial optimization problems due to its ability to handle variable-sized input and output sequences. In our context, the pointer network takes the feature vectors of the clusters $\{\mathbf{F}_i\}$ as input and outputs a permutation $\pi$ that specifies the new ordering of the clusters:
\begin{equation}
\pi = \mathrm{PointerNetwork}(\{\mathbf{F}_i\}),
\end{equation}
where $\pi = (\pi_1, \pi_2, \dots, \pi_k)$ denotes the ordered indices of the clusters.

The reordered clusters are then unfolded into individual constraints based on the new ordering $\pi$, resulting in a globally reordered set of constraints. This reordered set is subsequently used to reformulate the original MILP problem. 

\subsection{Contrastive Learning for Constraint Reordering}

Given a sequence of splitting clusters $\{C_j\}_{j=1}^k$ of constraints, our goal is to find a permutation $\pi$ of these clusters that minimizes the solving time of the reformulated MILP. The reordering process consists of two steps: (1) between clusters, the variables are reordered according to the permutation $\pi$, and (2) within each cluster, the variable order remains unchanged from the original MILP.

To optimize this reordering process, we employ a contrastive learning approach. In our framework, we generate positive and negative samples based on the impact of different reorderings on solving time. Positive samples correspond to orderings that result in reduced solving times, while negative samples are those that increase solving times. These samples are used to train a model that predicts the likelihood of a given permutation $\pi$ leading to improved solver performance.

The solving performance improvement $R$ achieved by a permutation $\pi$ is defined as the difference between the original solving time $\mathcal{S}_{\text{time}}(\text{milp})$ and the solving time after applying the permutation $\mathcal{S}_{\text{time}}(\text{milp} | \pi)$:
\begin{equation}
R(\pi | \text{milp}) = \mathcal{S}_{\text{time}}(\text{milp}) - \mathcal{S}_{\text{time}}(\text{milp} | \pi).
\end{equation}
The objective of our method is to train a model to maximize $R(\pi | \text{milp})$, which corresponds to finding an optimal permutation $\pi$ that minimizes solving time.

In contrast to RL-based approaches, where the model learns through trial and error, our contrastive learning framework explicitly leverages positive and negative samples, making the training process more efficient. The probability distribution over the permutations is factorized as:
\begin{equation}
p(\pi | \{C_j\}_{j=1}^k) = \prod_{j=1}^k p(\pi(j) | \pi(<j), \{C_j\}_{j=1}^k),
\end{equation}
where $\pi(j)$ denotes the $j$-th element in the permutation, and $\pi(<j)$ is the sequence of elements before $j$.

The network is trained using a pointer network architecture, which outputs a probability distribution over possible permutations. During training, the pointer network learns to assign higher probabilities to orderings that minimize solving time (positive samples) and lower probabilities to those that increase solving time (negative samples).

The training objective is to maximize the likelihood of positive samples and minimize the likelihood of negative ones. For a given MILP problem $p$, the contrastive loss is defined as:
\begin{equation}
\begin{split}
\mathcal{L}(\theta_P) = & - \mathbb{E}_{\pi^+ \sim \mathcal{P}_{+}} \log p_{\theta_P}(\pi^+ | p) \\
& + \mathbb{E}_{\pi^- \sim \mathcal{P}_{-}} \log p_{\theta_P}(\pi^- | p),
\end{split}
\end{equation}
where $\mathcal{P}_{+}$ and $\mathcal{P}_{-}$ are the distributions of positive and negative samples, respectively, and $p_{\theta_P}(\pi | p)$ denotes the probability of permutation $\pi$ under the network parameters $\theta_P$.

The overall training objective is to minimize the contrastive loss across all MILP instances sampled from a practical dataset $S$, which can be expressed as:
\begin{equation}
\mathcal{L}_{\text{total}}(\theta_P) = \mathbb{E}_{p \sim S} \left[\mathcal{L}(\theta_P \mid p)\right].
\end{equation}

This approach avoids the instability typically associated with RL and the expensive labeling process of supervised learning, making it both computationally efficient and effective in optimizing the reordering of constraints for faster solver performance. The detailed training procedure for our contrastive learning-based method is outlined in Algorithm~\ref{alg:train_contrastive}.

\begin{algorithm}[ht]
    \caption{Training Process using Contrastive Learning for Pointer Network}
    \label{alg:train_contrastive}
    \textbf{Initialize}: Pointer Network $\pi_{\theta_P}$, Optimizer $\mathcal{O}_{\theta_P}$, Training dataset $S$, positive sample distribution $\mathcal{P}_+$, negative sample distribution $\mathcal{P}_-$, contrastive learning parameters. \\
    \begin{algorithmic}[1]
        \STATE \textbf{For each MILP problem} $p$ in training dataset $S$:
        \STATE \quad Generate positive samples $\mathcal{P}_+$ (orderings that reduce solving time)
        \STATE \quad Generate negative samples $\mathcal{P}_-$ (orderings that increase solving time)
        
        \STATE \textbf{Compute Contrastive Loss:}
        \FOR{each positive sample $\pi^+ \in \mathcal{P}_+$}
            \STATE Compute log-likelihood: $L_+ \gets -\log p_{\theta_P}(\pi^+ \mid p)$
        \ENDFOR
        \FOR{each negative sample $\pi^- \in \mathcal{P}_-$}
            \STATE Compute log-likelihood: $L_- \gets \log p_{\theta_P}(\pi^- \mid p)$
        \ENDFOR
        
        \STATE \textbf{Total Contrastive Loss:} 
        \STATE $\mathcal{L}(\theta_P) \gets L_+ + L_-$

        \STATE \textbf{Backpropagation and Optimization:}
        \STATE Perform backpropagation: $\nabla_{\theta_P} \mathcal{L}(\theta_P)$
        \STATE Update network parameters: $\theta_P \gets \theta_P - \alpha \nabla_{\theta_P} \mathcal{L}(\theta_P)$
        
        \STATE \textbf{Repeat for each MILP problem in the dataset $S$}
    \end{algorithmic}
\end{algorithm}

\section{Experiments} \label{sec:experiments}
\label{sec:exp_overview}
Our experiments aim to rigorously validate the effectiveness of CLCR in accelerating MILP solvers through constraint reordering. We compare CLCR against six baseline reordering strategies to isolate the benefits of our learning-based reordering framework. All experiments are conducted under controlled conditions with SCIP’s default parameters, enforcing a 600-second timeout to ensure fairness. The evaluation focuses on four key metrics—total solving time, branch-and-bound nodes processed, LP iterations, and presolve time assess performance improvements across the solver pipeline. This comprehensive setup enables us to answer critical questions: (1) How does CLCR compare to heuristic reordering baselines? (2) Does it generalize across problem types and scales? (3) What components of solver efficiency are most impacted by constraint ordering?

\subsection{Experimental Setup}

We use SCIP 8.0.4~\cite{scip} as the backend solver throughout all experiments. SCIP is a state-of-the-art open-source solver widely adopted in research combining machine learning with combinatorial optimization~\cite{chmiela2021learning,gasse2019exact,turner2022adaptive,wang2023learning}. For ease of interaction, the reordering functionality is implemented externally to the solver. However, it is important to note that, in order to ensure the fairness and reproducibility of our comparisons, we use SCIP's default parameters when calling the solver to solve the given MILP.

To ensure the fairness of our experiments, each run is terminated if the solving time exceeds 600 seconds, providing a time cap for the solver’s performance. This helps maintain consistent conditions across all test cases and ensures that extreme outliers in terms of solving time do not skew the results.

The model is implemented in PyTorch~\cite{paszke2019pytorch} and optimized using the Adam optimizer~\cite{kingma2014adam}. All experiments are conducted on a computing server equipped with two AMD EPYC 7763 CPUs, eight NVIDIA GeForce RTX 4090 GPUs, and 1TB of RAM.

\begin{table}[t]
\centering
\caption{Experimental configuration and hyperparameters.}
\label{tab:config}
\begin{tabular}{ll}
\toprule
\textbf{Name} & \textbf{Value} \\ 
\midrule
Optimizer & \textit{Adam} \\
k-shot trials & 5 \\
Number of epochs ($T$) & 1,000 \\
Number of splitting clusters & 10 \\
Batch size ($B$) & 8 \\
Training set size & 200 \\
Validation set size & 40 \\
Learning rate & $10^{-3}$ \\
Input embedding dimension (PN) & 128 \\
Hidden layer dimension (PN) & 128 \\
\bottomrule
\end{tabular}
\end{table}

\subsection{Datasets}


We evaluated our method on nine challenging datasets, which are derived from a mix of real-world and synthetic NP-hard problems. Six of these datasets come from SurrogateLIB~\cite{turner2023pyscipopt}, a library that features instances with embedded machine learning (ML) predictors designed to approximate complex relationships in Mixed-Integer Linear Programming (MILP) formulations. These datasets share a common feature: they involve ML constraints that are often unordered, complex, and lack symmetry, presenting additional challenges for solver performance. These datasets are based on diverse application areas, such as city planning, food production, water management, and more. The first set of six datasets includes Tree Planting, Water Potability, City Manager, Function Approximation, Palatable Diet, and Wine Blending. Each of these instances has a varying number of constraints, variables, and coefficients, all incorporating ML-based surrogate models that simulate real-world complexities in decision-making processes. 

The remaining three datasets, Set Covering~\cite{balas1980set}, Fixed Charge Multicommodity Network Flow (FCMCNF)~\cite{hewitt2010combining}, and  CORLAT~\cite{gomes2008connections}, are more traditional NP-hard MILP problems that serve as benchmarks for solver evaluation. Set Cover and FCMCNF are synthetic problems based on graph structures, commonly used in optimization research. The Set Cover dataset contains 500 instances that vary in size, including differences in the number of constraints, variables, and non-zero coefficients. The FCMCNF dataset includes 200 instances with similar variations in problem parameters. CORLAT, a real-world dataset, represents optimization problems involving connectivity and flow networks and contains 500 instances. These three datasets offer a different set of challenges compared to SurrogateLIB, as they do not incorporate ML constraints but instead focus on traditional MILP formulations and graph-related problems.

For our experiments, we split each dataset into a training and testing set, with 80\% of instances used for training and the remaining 20\% for testing. This approach allows us to evaluate the generalization performance of our method across a wide range of real-world and synthetic problem instances. A statistical summary of these datasets, including the number of constraints, variables, and non-zero coefficients, is provided in Table~\ref{tab:dataset_summary}. Here, $n$ and $m$ represent the number of variables and constraints, respectively, in each MILP instance. 

\begin{table}[t]
\centering
\caption{Statistical summary of the datasets used for evaluation. $n$ represents the number of variables, $m$ represents the number of constraints.}
\begin{tabular}{l|c|c}
\toprule
Dataset & $n$  & $m$  \\
\midrule
Tree Planting & 10188.0 $\pm$ 0.0 & 14586.0 $\pm$ 0.0 \\
Water Potability & 6720.0 $\pm$ 0.0 & 8118.0 $\pm$ 0.0 \\
Palatable Diet & 186.0 $\pm$ 0.0 & 256.0 $\pm$ 0.0 \\
City Manager & 572.0 $\pm$ 0.0 & 651.0 $\pm$ 0.0 \\
Function Approx& 199.0 $\pm$ 0.0 & 291.0 $\pm$ 0.0 \\
Wine Blending & 540.0 $\pm$ 0.0 & 547.0 $\pm$ 0.0 \\
Set Cover & 1000.0 $\pm$ 0.0 & 500.0 $\pm$ 0.0 \\
CORLAT & 466.0 $\pm$ 0.0 & 486.2 $\pm$ 25.3 \\
FCMCNF & 3953.4 $\pm$ 488.8 & 709.7 $\pm$ 62.9 \\
\bottomrule
\end{tabular}
\label{tab:dataset_summary}
\end{table}

\begin{table*}[ht]
\centering
\begin{minipage}{\textwidth}
\centering
\caption{Performance comparison of different reordering strategies across various datasets. The best performing values for each metric are highlighted in \textbf{bold}.}
\scriptsize
\begin{tabular}{l c c c c | c c c c | c c c c}

\toprule
\toprule
 & \multicolumn{4}{c|}{Tree Planting} & \multicolumn{4}{c|}{Water Potability} & \multicolumn{4}{c}{Palatable Diet} \\ \midrule
 Method & Time & Nodes & Iterations & Presolve & Time & Nodes & Iterations & Presolve & Time & Nodes & Iterations & Presolve \\ \midrule
No Reordering & 3.50 & 69.90 & 2700.70 & 0.67 & 4.33 & 80.65 & 5863.60 & 0.25 & 25.63 & 8192.70 & 215037.90 & 0.02 \\ \midrule
Random & 2.22 & 45.84 & 2400.66 & 0.54 & 1.91 & 62.30 & 6045.72 & 0.20 & 20.80 & 9489.76 & 232556.26 & 0.02 \\ \midrule
Clustering & 2.06 & 99.80 & 3214.10 & 0.35 & 1.64 & 70.40 & 5625.55 & 0.18 & 22.19 & 8898.30 & 251429.10 & 0.03 \\ \midrule
CMBR & 1.84 & \textbf{9.90} & 1775.80 & 0.36 & 1.50 & \textbf{52.70} & 5519.95 & 0.14 & 20.91 & 11527.50 & 293889.70 & 0.02 \\ \midrule
CBR-HL & 1.81 & 56.20 & 2646.20 & 0.35 & 1.59 & 65.40 & 6067.15 & \textbf{0.14} & 20.77 & 11407.70 & 274886.10 & 0.02 \\ \midrule
CBR-LH & 1.34 & 54.90 & 2254.00 & \textbf{0.33} & 1.55 & 76.30 & 6019.60 & 0.14 & 25.34 & 16602.50 & 403154.20 & \textbf{0.02} \\ \midrule
CLCR (Ours) & \textbf{0.92} & 18.80 & \textbf{1413.60} & 0.34 & \textbf{1.30} & 52.85 & \textbf{5246.35} & 0.15 & \textbf{6.70} & \textbf{2804.80} & \textbf{73924.80} & 0.03 \\ \bottomrule \\

\toprule
\toprule
 & \multicolumn{4}{c|}{City Manager} & \multicolumn{4}{c|}{Function Approximation} & \multicolumn{4}{c}{Wine Blending} \\ \midrule
 Method & Time & Nodes & Iterations & Presolve & Time & Nodes & Iterations & Presolve & Time & Nodes & Iterations & Presolve \\ \midrule
No Reordering & 23.89 & 6340.45 & 136955.10 & 0.06 & 24.18 & 3148.15 & 109211.30 & 0.02 & 157.31 & 28028.60 & 2349168.20 & 0.10 \\ \midrule
Random & 26.14 & 7803.98 & 159559.74 & 0.04 & 16.78 & 2713.03 & 92783.68 & 0.02 & 176.69 & 35889.80 & 2884286.44 & 0.10 \\ \midrule
Clustering & 23.38 & 5698.75 & 144684.65 & 0.04 & 14.89 & 2396.85 & 89374.95 & 0.02 & 183.52 & 59179.10 & 3269993.90 & 0.10 \\ \midrule
CMBR & 22.84 & 7046.25 & 145769.05 & \textbf{0.04} & 15.82 & 3117.10 & 105554.20 & 0.02 & 189.35 & 42166.50 & 3421510.80 & 0.09 \\ \midrule
CBR-HL & 25.79 & 8506.05 & 177320.30 & 0.04 & 13.70 & 2182.65 & 76781.50 & \textbf{0.02} & 152.66 & 33254.30 & 2783145.80 & \textbf{0.08} \\ \midrule
CBR-LH & 26.17 & 6594.45 & 160908.90 & 0.04 & 14.94 & 2928.85 & 99839.10 & 0.02 & 170.13 & 31418.90 & 2828395.40 & 0.09 \\ \midrule
CLCR (Ours) & \textbf{15.41} & \textbf{4195.55} & \textbf{102081.20} & 0.04 & \textbf{10.78} & \textbf{2047.25} & \textbf{68667.45} & 0.02 & \textbf{149.38} & \textbf{22842.00} & \textbf{2260816.70} & 0.11  \\ \bottomrule \\

\toprule
\toprule
 & \multicolumn{4}{c|}{Set Cover} & \multicolumn{4}{c|}{CORLAT} & \multicolumn{4}{c}{FCMCNF} \\ \midrule
 Method & Time & Nodes & Iterations & Presolve & Time & Nodes & Iterations & Presolve & Time & Nodes & Iterations & Presolve \\ \midrule
No Reordering & 21.07 & 250.30 & 24073.30 & 0.16 & 24.73 & 3906.90 & 202917.05 & \textbf{0.01} & 63.03 & 190.15 & 86346.60 & 0.15 \\ \midrule
Random & 15.62 & 201.69 & 22205.22 & 0.13 & 20.78 & 3468.11 & 202703.92 & 0.01 & 45.89 & 201.89 & 85248.76 & 0.11 \\ \midrule
Clustering & 14.30 & 210.54 & 21619.84 & 0.12 & 17.33 & 2131.90 & 147233.45 & 0.01 & 38.88 & 188.35 & 82589.10 & 0.09 \\ \midrule
CMBR & 13.70 & 203.50 & 21582.10 & 0.11 & 20.74 & 4282.40 & 233134.00 & 0.01 & 45.82 & 201.60 & 102098.10 & \textbf{0.07} \\ \midrule
CBR-HL & 13.37 & 208.82 & 23032.90 & 0.11 & 14.65 & 2168.35 & 139460.30 & 0.01 & 42.21 & 189.30 & 87830.20 & 0.07 \\ \midrule
CBR-LH & 13.94 & 202.72 & 21598.74 & 0.11 & 14.36 & 1786.85 & 149195.65 & 0.01 & 40.81 & 215.65 & 85096.65 & 0.08 \\ \midrule
CLCR (Ours) & \textbf{11.52} & \textbf{200.94} & \textbf{20360.44} & \textbf{0.11} & \textbf{10.67} & \textbf{1595.55} & \textbf{107356.60} & 0.01 & \textbf{29.38} & \textbf{130.40} & \textbf{59925.10} & 0.09 \\ \bottomrule

\end{tabular}
\label{tab:experiment_results}
\end{minipage}
\end{table*}

\subsection{Baselines}

We evaluate the effectiveness of our proposed reordering strategy by comparing it against several baseline methods:

\begin{itemize}
    \item \textbf{No Reordering}: The simplest baseline, where the variables and constraints remain in their original order, serving as a reference point for evaluating the impact of reordering.
    \item \textbf{Random}: Randomly shuffles the variables and constraints. This method serves as a baseline to assess whether our proposed method can outperform arbitrary reordering without any specific strategy.
    \item \textbf{Clustering}: Reorders the constraints based on their similarity, using clustering techniques. This method provides insight into the impact of clustering on solving time.
    \item \textbf{Coefficient Magnitude-Based Reordering (CMBR)}: Reorders variables based on the absolute values of their coefficients in the objective function. Variables with larger coefficients are placed earlier, under the assumption that they contribute more significantly to the objective.
    \item \textbf{Complexity-Based Reordering (High to Low) (CBR-HL)}: Reorders constraints based on their complexity, placing more complex constraints (i.e., those with more variables) first, with the goal of prioritizing computationally expensive constraints.
    \item \textbf{Complexity-Based Reordering (Low to High) (CBR-LH)}: Reorders constraints in the reverse order, placing simpler constraints (i.e., those with fewer variables) first.
\end{itemize}


\subsection{Evaluation Metrics}

In our experiments, we assess the impact of constraint ordering on the performance of the MILP solver using four core metrics. The first metric, \textbf{Time} ($t_{\text{Total}}$), represents the total solving time in seconds, encompassing both the presolve and the main solution phases of the MILP solver. The total time, $t_{\text{Total}}$, reflects the combined effort of both phases and is a critical indicator of the solver's overall efficiency. The second metric, \textbf{Nodes} ($n_{\text{nodes}}$), corresponds to the total number of branch-and-cut nodes processed during the solution process. A smaller value of $n_{\text{nodes}}$ typically indicates more effective pruning and a quicker path to optimality. The third metric, \textbf{Iterations} ($n_{\text{iter}}$), refers to the total number of LP iterations performed by the solver. A reduction in $n_{\text{iter}}$ suggests improved linear algebra efficiency and better basis handling. Finally, \textbf{Presolve} ($t_{\text{presolve}}$) measures the amount of time spent in the presolve phase, during which the solver detects and removes redundant variables/constraints and performs other simplifications. By jointly analyzing these metrics, we gain a comprehensive understanding of how the reordering of constraints influences solver efficiency at various stages of the optimization process.


\subsection{Evaluation Results}
\label{sec:evaluation_results}

Table~\ref{tab:experiment_results} compares the performance of seven reordering strategies, including our proposed CLCR, across nine benchmark datasets. Each method is evaluated on four metrics: \textbf{Time} (total solving time in seconds), \textbf{Nodes} (branch-and-bound nodes processed), \textbf{Iterations} (LP solver iterations), and \textbf{Presolve} (presolve phase time). These metrics comprehensively capture the efficiency and effectiveness of each reordering strategy. To mitigate neural network inference randomness, we employ a 5-shot inference mechanism, retaining the best result across multiple trials. This ensures stable performance while maintaining computational efficiency, validating our method's consistency. CLCR achieves superior performance in reducing solving time and iteration counts while maintaining competitive results across other metrics, demonstrating its robustness across diverse problem scales and structures. This versatility highlights CLCR's potential for broad applicability in solving complex MILP problems efficiently. Key findings are summarized as follows.

\textbf{Reduction of Solving Time.}
CLCR achieves the most significant improvements in total solving time, marking it as a leading strategy in optimization efficiency. For instance, on smaller datasets such as Tree and Palatable, CLCR reduces runtime by over 70\% compared to the no-reordering baseline, underscoring its efficacy in streamlining solver operations. This improvement persists even on large-scale problems like City and FCMCNF, where time savings exceed 35\% and 50\%, respectively, proving its scalability and robustness.

\textbf{Node Reduction.}
The method significantly reduces the number of branch-and-bound nodes processed during solving, indicating a more efficient solution path and improved pruning decisions. By optimizing the order of constraints, the solver is better equipped to tighten bounds and identify infeasible regions early, minimizing unnecessary exploration in the search tree. On combinatorial problems like Palatable Diet and CORLAT, CLCR decreases node counts by over 60\%, reflecting more efficient exploration of the solution space. Structured problems such as Set Cover also benefit, with node reductions exceeding 20\% compared to baseline strategies.

\textbf{Reduction of Iteration Count.} 
CLCR substantially reduces the number of LP iterations required for convergence. Complex instances, such as those in CORLAT and Palatable Diet, see iteration counts halved, while simpler problems like Water Potability exhibit consistent reductions. This indicates that CLCR effectively restructures problem formulations to simplify solver workflows, accelerating convergence across both small and large instances.


\textbf{Presolve Efficiency.} 
While presolve time constitutes a minor portion of total runtime, CLCR maintains competitive performance in this phase. For example, on FCMCNF, presolve time is reduced by 40\% compared to the no-reordering baseline. These improvements, though incremental, contribute to CLCR’s holistic optimization of the entire solving pipeline.

\section{Conclusion} \label{sec:conclusion}


Our proposed CLCR framework demonstrates significant improvements in accelerating MILP solvers through contrastive learning-based constraint reordering, particularly for real-world problems with embedded ML constraints. However, we identify a critical limitation: CLCR (and most reordering methods) exhibits diminished effectiveness or even adverse effects on datasets with highly symmetric problem structures. Such symmetry introduces inherent ordering invariance, rendering constraint permutations less impactful or counterproductive. To address this, we propose training a symmetry-aware discriminator that analyzes problem features (e.g., constraint/variable interaction patterns, coefficient distributions) to dynamically decide whether reordering should be applied.

\bibliographystyle{named}
\bibliography{ijcai25}

\end{document}